\documentclass{article}

\usepackage{microtype}
\usepackage{graphicx}
\usepackage{subcaption}
\usepackage{booktabs} 

\usepackage{graphicx}
\usepackage{todonotes}
\usepackage{amsmath}
\usepackage{amssymb}
\usepackage{pifont}
\newcommand{\cmark}{\ding{51}}%
\newcommand{\xmark}{\ding{55}}%
\newcommand{\omark}{\ding{58}}%
\usepackage{bbm}
\usepackage{booktabs}
\usepackage{url}
\usepackage{subcaption}

\usepackage{breakurl}
\usepackage[breaklinks]{hyperref}

\usepackage{makecell}
\usepackage{multicol}
\usepackage{multirow}

\newcommand{\algo}[1]{{\sc #1}}
\newcommand{\quickscorer}{\algo{QuickScorer}}
\newcommand{\vquickscorer}{\algo{V-QuickScorer}}
\newcommand{\rapidscorer}{\algo{RapidScorer}}

\usepackage{setspace}

\usepackage[accepted]{mlsys2023}

\mlsystitlerunning{Fast Inference of Tree Ensembles on ARM Devices}

\begin{document}
	
	\twocolumn[
	\mlsystitle{Fast Inference of Tree Ensembles on ARM Devices}
	
	
	
	\mlsyssetsymbol{equal}{*}
	
	\begin{mlsysauthorlist}
		\mlsysauthor{Simon Koschel}{tu}
		\mlsysauthor{Sebastian Buschjäger}{tu}
		\mlsysauthor{Claudio Lucchese}{up}
		\mlsysauthor{Katharina Morik}{tu}
	\end{mlsysauthorlist}
	
	\mlsysaffiliation{tu}{Artificial Intelligence Group, TU Dortmund, Germany}
	\mlsysaffiliation{up}{Ca’ Foscari Univ. of Venice, Italy}
	
	\mlsyscorrespondingauthor{Simon Koschel}{simon.koschel@tu-dortmund.de}
	\mlsyscorrespondingauthor{Sebastian Buschjäger}{sebastian.buschjaeger@tu-dortmund.de}
	
	\mlsyskeywords{Machine Learning, MLSys}
	
	\vskip 0.3in
	
	\begin{abstract}
		With the ongoing integration of Machine Learning models into everyday life, e.g. in the form of the Internet of Things (IoT), the evaluation of learned models becomes more and more an important issue. Tree ensembles are one of the best black-box classifiers available and routinely outperform more complex classifiers. While the fast application of tree ensembles has already been studied in the literature for Intel CPUs, they have not yet been studied in the context of ARM CPUs which are more dominant for IoT applications.
		In this paper, we convert the popular {\quickscorer} algorithm and its siblings from Intel's AVX to ARM's NEON instruction set. Second, we extend our implementation from ranking models to classification models such as Random Forests. Third, we investigate the effects of using fixed-point quantization in Random Forests.
		Our study shows that a careful implementation of tree traversal on ARM CPUs leads to a speed-up of up to $9.4$ compared to a reference implementation. Moreover, quantized models seem to outperform models using floating-point values in terms of speed in almost all cases, with a neglectable impact on the predictive performance of the model. Finally, our study highlights architectural differences between ARM and Intel CPUs and between different ARM devices that imply that the best implementation depends on both the specific forest as well as the specific device used for deployment.
	\end{abstract}
	]
	
	
	
	\printAffiliationsAndNotice{}  

	\section{Introduction}
	\label{sec:intro}
	
	Ensemble algorithms offer state-of-the-art performance in many applications and often outperform single classifiers by a large margin. With the ongoing integration of embedded systems and machine learning models into our everyday life, e.g., in the form of the Internet of Things (IoT), the hardware platforms that execute ensembles must also be considered. Common IoT applications such as pedestrian detection \cite{Marin/etal/2013}, 3D face analysis \cite{Fanelli/etal/2013}, noise signal analysis \cite{Saki/etal/2016} or nano-particle analysis \cite{yayla/etal/2019} require real-time predictions over a continuous stream of measurements. Hence, large pre-trained ML models must be deployed to and executed on IoT devices directly. While there is a large, heterogeneous landscape of different edge devices available, the vast majority of these devices offer comparably small resources (e.g., few kilobytes of memory and limited processing speed) while using only a fraction of the power required for conventional desktop systems. Table \ref{tab:mcus} gives an overview of common micro-controller units (MCUs) and their properties. Most IoT devices either use an ARM computing architecture or an ATMega architecture in contrast to the x86 architecture most commonly found in laptop and desktop hardware. To deploy tree ensembles onto these devices, an efficient implementation of the tree traversal must be given. 
	
	\begin{table*}[h]
		\caption[Typical microcontroller units (MCUs) found in edge and IoT devices.]{Typical microcontroller units (MCUs) found in edge and IoT devices. The top group shows bare-metal MCUs, which typically do not run an operating system. The bottom group shows MCUs that typically also run an operating system. \cmark~ denotes the availability of a feature, \xmark~ marks its absence, and \omark~ denotes optional/partial support. The original table is due to \cite{branco/etal/2019}, but the float, SIMD, word, and architecture columns have been added by taking the corresponding values from the corresponding data sheets. For comparison, the Intel i7-7700K CPU has also been added as a typical desktop/server CPU.}
		\label{tab:mcus}
		\centering
		\begin{tabular}{lrrrrrrrr}
			\hline
			MCU     & Arch.                    & Clock     & Float    & Word & SIMD    & Flash   & (S)RAM    & Power         \\ \hline
			Arduino Uno & ATMega128P    & 16MHz     & \xmark       & 8bit     & \xmark      & 32KB    & 2KB       & 12mA          \\
			Arduino Mega & ATMega2560   & 16MHz     & \xmark       & 8bit     & \xmark      & 256KB   & 8KB       & 6mA           \\
			Arduino Nano &ATMega2560   & 16MHz     & \xmark       & 8bit     & \xmark      & 26-32KB & 1-2KB     & 6mA           \\
			STM32L0 &ARM Cortex-M0         & 32MHz     & \xmark       & 32bit    & \xmark      & 192KB   & 20KB      & 7mA           \\
			Arduino MKR1000 & ARM Cortex-M0 & 48MHz     & \xmark       & 32bit    & \xmark      & 256KB   & 32KB      & 4mA           \\
			Arduino Due & ARM Cortex-M3     & 84MHz     & \xmark       & 32bit    & \xmark      & 512KB   & 96KB      & 50mA          \\
			STM32F2 & ARM Cortex-M3         & 120MHz    & \xmark       & 32bit    & \xmark      & 1MB     & 128KB     & 21mA          \\
			STM32F4 & ARM Cortex-M4         & 180MHz    & \omark & 32bit    & \omark & 2MB     & 384KB     & 50mA          \\ \hline
			Raspberry PI A+ & ARMv6                      & 700MHz    & \cmark      & 32bit    & \xmark      & SD Card & 256MB     & 80mA          \\
			Raspberry PI Zero & ARMv6                  & 1GHz      & \cmark      & 32bit    & \xmark      & SD Card & 512MB     & 80mA          \\
			Raspberry PI 3B  & ARMv8                    & 4@1.2GHz  & \cmark      & 64bit    & \cmark     & SD Card & 1GB       & 260mA         \\
			Intel i7-7700K & x86              & 4@4.5GHz & \cmark      & 64bit    & \cmark     & HDD/SSD & 2 - 64GB & $\approx$ 80A \\ \hline
		\end{tabular}
	\end{table*}
	
	{\quickscorer} is an efficient tree traversal algorithm that discards the tree structure entirely but executes comparisons across multiple trees inside a forest by adopting a novel bitvector representation of the forest. 
	It received multiple extensions including vectorization through SIMD, a GPU version and a novel compression scheme to enable a more efficient traversal of larger trees \cite{lucchese/etal/2015,lucchese/etal/2016,lucchese/etal/2018,ye/etal/2018}. Unfortunately, these variations have not been adopted for the ARM architecture but are presented for x86 and CUDA exclusively. While there is an overlap between the x86 and ARM architecture, we find significant differences in their handling of vectorization instructions and floating point instructions. More specifically, modern x86 CPUs usually offer SIMD registers with 128 to 512 bit (e.g., AVX2 or AVX-512), whereas ARM's SIMD instruction set NEON only allows up to 128 bit words. Moreover, x86 has rich support for floating point values, whereas ARM has limited support for floating point handling. To enable the execution of large ensembles on small devices, we, therefore, discuss how to efficiently adopt the {\quickscorer} algorithm family to the ARM architecture. Moreover, we investigate the effects of fixed-point quantization on tree traversal to mitigate the costly handling of floating points. Our contributions are as follows: 
	\begin{itemize}
		\item \textbf{ARM QuickScorer:} We present the first adaption of the {\quickscorer} algorithm family including {\quickscorer} \cite{lucchese/etal/2015}, {\vquickscorer} \cite{lucchese/etal/2016} and {\rapidscorer} \cite{ye/etal/2018} for ARM devices. Our adaption retains the execution speed of its Intel counterpart and enables the traversal of large additive tree ensembles on small devices. 
		\item \textbf{Evaluating quantization for trees:} We experimentally evaluate how the accuracy of tree ensembles changes when thresholds and leaf values are represented in (limited precision) fixed-point data types. We show that this approach offers consistent speed-ups compared to using floating points with minor impacts on the classification accuracy. 
		\item \textbf{Extensive comparison on different devices:} We present an extensive experimental study of state-of-the-art tree traversal algorithms on a variety of different ARM hardware platforms and datasets. We find that {\rapidscorer} as well as {\vquickscorer} generally perform best with some outliers in some cases.
	\end{itemize}
	
	This paper is organized as the following. Section \ref{sec:rel-work} gives an overview of related work and our notation. Section \ref{sec:QS} introduces the {\quickscorer} algorithms for Intel and section \ref{sec:QSonARM} details how to adapt them for the ARM architecture. Section \ref{sec:quant} discusses the quantization of leaves and splits in decision trees. Last, section \ref{sec:experiments} presents our experiments and section \ref{sec:conclusion} concludes the paper. 
	
	\section{Background and Notation}
	\label{sec:rel-work}
	
	We consider a supervised learning setting, in which we assume that training and test points are drawn i.i.d. according to some distribution $\mathcal D$ over the input space $\mathcal X$ and labels $\mathcal Y$. For training, we have given a labeled sample $\mathcal{S} = \{(x_i,y_i)|i=1,\dots,N\}$, where $x_i \in \mathcal X \subseteq \mathbb R^d$ is a $d$-dimensional feature-vector and $y_i\in \mathcal Y \subseteq \mathbb R^C$ is the corresponding target vector. For regression and ranking problems we have $C=1$ and $\mathcal Y = \mathbb R$. For classification problems with $C \ge 2$ classes we encode each label as a one-hot vector $y = (0,\dots,0,1,0,\dots,0)$ which contains a `$1$' at coordinate $c$ for label $c \in \{0,\dots,C-1\}$. We assume that we have given a pre-trained axis-aligned decision tree ensemble with $M$ classifiers $h_i \in \mathcal H$ of the following form:
	\begin{equation}
		\label{eq:ensemble}
		f(x) = \sum_{i=1}^M h_i(x)
	\end{equation}
	If the ensemble is weighted (e.g., as in AdaBoost) using weights $w_i$ or a majority vote is used, i.e., $w_i = \frac{1}{M}$ (e.g., as in Random Forests), then we re-scale the individual classifier's predictions to include the weight:
	$$
	f(x) = \sum_{i=1}^M w_i {h'}_i(x) = \sum_{i=1}^M h_i(x)
	$$
	An axis-aligned decision tree (DT) partitions the input space $\mathcal X$ into increasingly smaller d-dimensional hypercubes called leaves and uses independent predictions for each leaf in the tree. Inner nodes perform an axis-aligned split $1\{x_k \le t\}$ where $k$ is a feature index and $t$ is a split-threshold. Each node is associated with split function $s(x) \colon \mathcal X \to \{0,1\}$ that is `1' if $x$ belongs to the hypercube of that node and `0' if not. Let $L$ be the total number of leaf nodes in the tree and let $L_i = (n_1,n_2,\dots)$ be the nodes visited depending on the outcome of $1\{x_k \le t\}$ that lead to leaf $i$ with prediction $\widehat y_i$, then the prediction function of a tree is given by
	\begin{equation}
		h(x) = \sum_{i = 1}^L \widehat y_i s_i(x)  = \sum_{i=1}^L \widehat y_i \prod_{n \in L_i} s_{i,n}(x) 
	\end{equation}
	where $\widehat y_i \in \mathbb R^C$ is the (constant) prediction value per leaf and $s_{i,l}$ are the split functions of the individual hypercubes. In addition to the pre-trained model, we are also given a target hardware platform (e.g., a specific smartphone) and a test sample $X_{test}$, and our goal is to apply the given ensemble $f$ to the test sample $X_{test}$ as quickly as possible on the given hardware. The fast application of DT ensembles has been studied in the literature. We find that there are two orthogonal lines of research: 
	
	\textbf{Changing the model:} The first line of research studies the training algorithm of DT ensembles and tries to find accurate and hardware-friendly DT ensembles. For example, `classic' decision tree pruning algorithms (e.g., minimal cost complexity pruning or sample complexity pruning) already reduce the size of DTs while offering a better accuracy (c.f. \cite{Barros/etal/2015}). Similar, in the context of model compression (see e.g. \cite{choudhary/etal/2020} for an overview) specific models such as Bonsai \cite{kumar/etal/2017} or Decision Jungles \cite{shotton/etal/2013} aim to find smaller tree ensembles already during training. Last, ensemble pruning (see e.g., \cite{Buschjaeger/Morik/2021b,lucchese/etal/2018}) aims to select fewer members from the ensemble to improve accuracy while reducing memory consumption.
	
	\textbf{Optimizing the execution of a DT ensemble:} The second line of research studies the traversal of pre-trained tree ensembles without changing the model. Asadi et al. introduce in \cite{Asadi/etal/2014} different implementation schemes for pre-trained tree ensemble models in the context of learning-to-rank tasks: The first one uses a while-loop to iterate over individual nodes of the tree, whereas the second approach decomposes each tree into its individual if-else structure. For the first implementation, the authors also consider a continuous data layout (i.e., an array of \emph{structs}) to increase data locality but do not directly optimize each implementation. Buschjäger et al. investigate this idea more in-depth in a series of contributions \cite{Buschjaeger/Morik/2017b,Buschjaeger/2018a,Chen/etal/2022}. They view the execution of DTs as a series of Bernoulli experiments in which either the left or the right child of a node is visited. Based on the estimated Bernoulli probabilities, they then lay out DTs, so that cache hits are maximized. The resulting \emph{tree-framing} algorithm is applicable to the two different DT implementations and can offer a speed-up of up to $2-4$ on a variety of systems. Kim et al. present in \cite{kim/etal/2010} an implementation for binary search trees using vectorization units on Intel CPUs and compare their implementation against a GPU implementation. The authors provide insights into how to tailor the implementation to Intel CPUs by taking into account register sizes, cache sizes as well as page sizes. GPUs have also been investigated for the traversal of DT ensemble. The first work in this direction was due to Sharp in \cite{sharp/2008} showing how to encode DTs as 2D textures that are then processed by the GPU. A more recent re-implementation of this approach can be found in \cite{nakandala/etal/2020}. Here, the authors discuss multiple strategies that map the execution of a DT into tensor operations. The key insight is that mapping DT traversal to tensor operations usually leads to an increase in computation, but this increase is justified due to the availability of more efficient tensor libraries and tensor processing hardware. 
	Unfortunately, small embedded systems often lack powerful GPUs or hardware accelerators that process tensors, so these traversal methods are unsuitable for an embedded system. Finally, Lucchese et al. present another traversal algorithm called {\quickscorer} in \cite{lucchese/etal/2015}. This approach offers a cache-efficient way to evaluate ensembles as a whole and outperforms other algorithms.
	This traversal algorithm has since been extended to SIMD operations \cite{lucchese/etal/2016} and GPUs \cite{lettich/etal/2018}. In \cite{ye/etal/2018} the vectorized {\quickscorer} algorithm is further extended by introducing a more efficient data layout.
	
	Unfortunately, the available literature using SIMD extensions (\cite{lucchese/etal/2016} and \cite{ye/etal/2018}) focuses on the x86 Intel architecture, whereas the common computer architecture in the realm of embedded systems is ARM. Hence, in the next section, we explain the {\quickscorer} algorithm and its siblings in more detail. After that, we survey how to adapt it for the ARM architecture. 
	
	\section{QuickScorer and its siblings}
	\label{sec:QS}
	
	
	The {\quickscorer} algorithm~\cite{lucchese/etal/2015} introduces several novelties that allow fast and efficient evaluation of a forest of binary decision trees for ranking problems.
	The pseudocode of the algorithm is reported in Algorithm~\ref{alg:qs}.
	First, the traversal of the forest is conducted in a feature-wise fashion: the algorithm evaluates the nodes performing a test on features $0$, then feature $1$, and so on.
	Each node of the current feature $k$ is then evaluated, and a bitvector \texttt{leafidx[$h$]} for the corresponding tree $h$ is updated to mark leaves that will not be reached by the instance's traversal. This is done by applying a logical AND of the bitvector of each tree \texttt{leafidx[$h$]} and the node's bitvector \texttt{bitmasks[$n$]}.
	This process is implemented in lines 5 - 13 of algorithm \ref{alg:qs}. Eventually, \texttt{leafidx[$h$]} identifies the exit leaf, where the current instance will end up, and a lookup table is used to retrieve the tree's prediction (lines 15 - 20).
	Second, thanks to this reorganization of the computation, {\quickscorer}'s data structure is implemented as a set of arrays on which one simple linear scan and bitwise operations are conducted. 
	\begin{algorithm}[h]
		\footnotesize
		\caption{{\quickscorer}($x, \mathcal{T}$)}
		\label{alg:qs}
		\begin{algorithmic}[1]
			\FOR{$T_h \in \mathcal{T}$}
			\STATE \texttt{leafidx[$h$]} $\gets$ \texttt{{11...11}}
			\ENDFOR
			\STATE \COMMENT{\textbf{Mask Computation}}
			\FOR{$f_k \in \mathcal{F}$}
			\FOR{$(\gamma,h,n) \in N_k$ in ascending order}
			\IF{$x[k] > \gamma$}
			\STATE \texttt{leafidx[$h$]} $\gets$ \texttt{leafidx[$h$]} $\land$ \texttt{bitmasks[$n$]}
			\ELSE
			\STATE \texttt{break}
			\ENDIF
			\ENDFOR
			\ENDFOR
			\STATE \COMMENT{\textbf{Score Computation}}
			\STATE $ \texttt{score} \gets 0$	
			\FOR{$T_h \in \mathcal{T}$}
			\STATE $ j \gets$ index of leftmost bit set to 1 of \texttt{leafidx[$h$]}
			\STATE $ l \gets h \cdot L + j$
			\STATE $ \texttt{score} \gets \texttt{score}$ + \texttt{leafvalues[$l$]}
			\ENDFOR
		\end{algorithmic}
	\end{algorithm}

	On top of {\quickscorer} (QS), the same authors implemented an improved variant, named {\vquickscorer} (VQS)~\cite{lucchese/etal/2016}.
	It keeps the main structure of the {\quickscorer} algorithm but uses SIMD extensions to evaluate $v$ instances at the same time.
	In \cite{lucchese/etal/2016} the used instruction sets can fit 128 and 256 bits. For feature values, the data type \texttt{float}, which is represented by 32 bits, is used. Therefore, $v = 4$ and $v = 8$ instances can be processed in parallel.
	The pseudocode is presented in Algorithm~\ref{alg:vqs}.
	Like {\quickscorer}, the algorithm iterates featurewise over the nodes of each tree. Then, each node's value is compared against the instances' feature values (lines 9 - 11). Using the resulting mask of this comparison, the node's bitmask is conditionally applied to the instances' bitvectors (lines 12 - 16). In the end, the exit leaves of each tree and instance are calculated, and the corresponding value of the exit leaf is added to the score (lines 23 - 31).
	Note that $x_i$, \verb|leafidx|$_i$[$h$], $j_i$ and $l_i$ now use the subscript $i$ denoting which of the $v$ instances they describe.
	A right arrow above a symbol denotes a SIMD vector with multiple different elements belonging to the multiple instances that are evaluated in parallel.
	A left arrow above a symbol denotes a SIMD vector with multiple identical elements.

	
	
	\begin{algorithm}[h]
		\footnotesize
		\caption{\textsc{V-QuickScorer}}
		\label{alg:vqs}
		\begin{algorithmic}[1]
			\FOR{$T_h \in \mathcal{T}$}
			\FOR{$i$ \textbf{to} $v - 1$}
			\STATE \texttt{leafidx$_{i}$[$h$]} $\gets$ \texttt{{11...11}}
			\ENDFOR
			\ENDFOR
			\STATE \COMMENT{\textbf{Mask Computation}}
			\FOR{$f_k \in \mathcal{F}$}
			\FOR{$(\gamma,h,n) \in N_k$ in ascending order}
			\STATE $\overrightarrow{\gamma} \gets$ ($\gamma,...,\gamma$)
			\STATE $\overrightarrow{x} \gets$ ($x_{v-1}[k],...,x_0[k]$)
			\STATE $\overrightarrow{mask} \gets \overrightarrow{x} > \overrightarrow{\gamma}$
			\IF{$\overrightarrow{mask} \neq 0$}
			\STATE $ \overrightarrow{m} \gets$ (\texttt{bitmasks}$[n]$, ..., \texttt{bitmasks}$[n]$)
			\STATE $ \overrightarrow{b} \gets$ (\texttt{leafidx}$_{v-1}[h]$, ..., \texttt{leafidx}$_{0}[h]$)
			\STATE $ \overrightarrow{y} \gets$ $\overrightarrow{m} \land \overrightarrow{b}$
			\STATE \texttt{leafidx}$_{(v-1), ..., 0}[h] \xleftarrow{\overrightarrow{mask}} \overrightarrow{y}$
			\ELSE
			\STATE \texttt{break}
			\ENDIF
			\ENDFOR
			\ENDFOR
			\STATE \COMMENT{\textbf{Score Computation}}
			\STATE $ \overrightarrow{s} \gets (0,...,0)$
			\FOR{$T_h \in \mathcal{T}$}
			\FOR{$i$ \textbf{to} $v - 1$}
			\STATE $j_i \gets$ index of leftmost 1 bit of \texttt{leafidx$_{i}[h]$}
			\STATE $l_i \gets h \cdot L + j_i$
			\ENDFOR
			
			\STATE $ \overrightarrow{v} \gets$ (\texttt{leafvalues}$[l_{v-1}]$, ..., \texttt{leafvalues}$[l_0]$)
			\STATE $ \overrightarrow{s} \gets \overrightarrow{s} + \overrightarrow{v}$
			\ENDFOR
		\end{algorithmic}
	\end{algorithm}

	{\rapidscorer} (RS)~\cite{ye/etal/2018} is a further improvement of {\vquickscorer} that also uses SIMD extensions.
	The authors show that the data structure adopted by {\quickscorer} can be made more compact, resulting in a reduced memory footprint and reduced number of operations.
	They present three main improvements: 
	First, they introduce a more compact data structure called epitome to represent the bitmasks of each tree node.
	Second, a merging mechanism is introduced to avoid the evaluation of redundant splits. Due to the traversal strategy of the {\quickscorer} algorithm, equal nodes that use the same threshold for the same feature are processed directly after one another. Hence, it performs many consecutive redundant comparisons for the same threshold. RS merges these equivalent nodes and, therefore, only performs one comparison per node.
	Last, the data layout of the bitvector \verb|leafidx| that identifies the exit leaf is being transposed. This means that the bitvector is no longer stored continuously in memory. For the instances that are evaluated in parallel, the bitvectors are stored inside multiple SIMD registers. The first register contains the first bytes of each instance's bitvector, the second register contains the second bytes, and so on.
	This layout allows to efficiently apply bytewise operations on the bitvectors, and it is denoted by $\overrightarrow{\texttt{leafidx}}$.
	
	\section{Converting QuickScorer to ARM}
	\label{sec:QSonARM}
	We start by converting {\quickscorer} and its siblings from Intel's SSE/AVX instruction set to ARM'S NEON instruction. Then, we show how the algorithms can be adapted from only supporting learning-to-rank tasks (with a scalar output) to also supporting classification (with a vector output). 
	Since {\quickscorer} does not use SIMD extensions, we can use the same implementation on Intel and ARM processors. Hence, we discuss {\vquickscorer} and {\rapidscorer} in more detail.
	
	\subsection{From Intel AVX to ARM NEON}
	One major difference between AVX and NEON is the SIMD register size. AVX registers are 256 bits wide, while NEON registers can only store 128 bits. This means that using NEON, one can only process half of the instances compared to AVX. Similarly, for VQS, this means that it can only process four instances, and RS can evaluate 16 instances in parallel.
	
	\noindent\textbf{V-QuickScorer}:
	In order to convert the Intel implementation to the ARM platform, we need to replace AVX instructions with equivalent NEON instructions. For {\vquickscorer}, this can be done without a lot of effort since most AVX instructions required for VQS have direct equivalents in NEON.
	
	\noindent\textbf{RapidScorer}:
	Most AVX instructions used for the {\rapidscorer} algorithm can also be directly translated to NEON instructions. An exception is the process of finding the correct exit leaf. As mentioned above, {\rapidscorer} uses the  bytewise transposed layout $\overrightarrow{\texttt{leafidx}}$.
	Here, the bitvector of one instance is represented by a single column in this data structure. The process of finding the exit leaf for AVX and NEON is implemented in algorithms \ref{alg:rs:vectorized_findleafidx:avx} and \ref{alg:rs:vectorized_findleafidx:neon} and is done as follows:
	Essentially we need to find the first bit set to \texttt{1} in the data structure $\overrightarrow{\texttt{leafidx}}$. First, we need to iterate through the SIMD registers from top to bottom and check for the first elements that contain a \texttt{1} (AVX: line 1 - 4, NEON: line 1 - 3).
	
	We save the instances where a \texttt{1} was already found using a bitmask (line 2).
	In NEON, the required comparison against $0$ can be optimized with the NEON intrinsic \verb|vtstq_u8| by using a vector that contains only ones. That already includes the negation of this comparison.
	We extract the current byte and put it into $\overrightarrow{b}$ (AVX: line 5, NEON: line 4).
	The index of this byte is put into $\overrightarrow{c_1}$ (AVX: line 6, NEON: line 5).
	Then we extract the position of the first \texttt{1} in $\overrightarrow{b}$ and save it into $\overrightarrow{c_2}$ (AVX: line 8, NEON: line 7).
	Therefore, the trailing zeros for 8 bit elements need to be counted. With AVX this can be implemented using a lookup table with \verb|_mm256_shuffle_epi8|. For NEON this is done by first reversing the vector with \verb|vrbitq_u8| and then counting the leading zeros with \verb|vclzq_u8|.
	Last, we multiply $\overrightarrow{c_1}$ by $8$ and add $\overrightarrow{c_2}$ (AVX: line 9 - 10, NEON: line 8).
	Therefore we use \verb|_mm256_slli_epi32| and \verb|_mm256_add_epi8| for AVX and \verb|vmlaq_u8| for NEON.
	

	\begin{algorithm}
		\footnotesize
		\caption{\textsc{Vectorized\_FindLeafindex} (AVX)}
		\label{alg:rs:vectorized_findleafidx:avx}
		\begin{algorithmic}[1]
			\FOR{$m = 0$ to $\lceil L / 8\rceil$}
			\STATE $\overrightarrow{y} \gets \neg \verb|_mm256_cmpeq_epi8|(\overrightarrow{\texttt{leafidx}[m]}, \overleftarrow{0})$
			\STATE $\overrightarrow{z} \gets \verb|_mm256_cmpeq_epi8|(\overrightarrow{b}, \overleftarrow{0})$
			\STATE $\overrightarrow{z} \gets \verb|_mm256_and_si256|(\overrightarrow{y}, \overrightarrow{z})$
			\STATE $\overrightarrow{b} \gets$ \verb|_mm256_blendv_epi8|($\overrightarrow{b}, \overrightarrow{\texttt{leafidx}[m]}, \overrightarrow{z})$
			\STATE $\overrightarrow{c_1} \gets \verb|_mm256_blendv_epi8|(\overrightarrow{c_1}, \overleftarrow{m}, \overrightarrow{z})$
			\ENDFOR
			\STATE $\overrightarrow{c_2} \gets \verb|_mm256_ctz_epi8|\footnotemark(\overrightarrow{b})$
			\STATE $\overrightarrow{c} \gets \verb|_mm256_slli_epi32|(\overrightarrow{c_1}, 3)$
			\STATE \textbf{return} $\verb|_mm256_add_epi8|(\overrightarrow{c_2}, \overrightarrow{c})$
		\end{algorithmic}
	\end{algorithm}
	\footnotetext{This intrinsic does not exist for AVX. It is implemented using a table lookup with \texttt{\_mm256\_shuffle\_epi8}.}
	
	\begin{algorithm}
		\footnotesize
		\caption{\textsc{Vectorized\_FindLeafindex} (NEON)}
		\label{alg:rs:vectorized_findleafidx:neon}
		\begin{algorithmic}[1]
			\FOR{$m = 0$ to $\lceil L / 8\rceil$}
			\STATE  $\overrightarrow{y} \gets \verb|vtstq_u8|(\overrightarrow{\texttt{leafidx}[m]}, \overleftarrow{1})$
			\STATE $\overrightarrow{z} \gets \verb|vandq_u8|(\overrightarrow{y}, \verb|vceqq_u8|(\overrightarrow{b}, \overleftarrow{0}))$
			\STATE $\overrightarrow{b} \gets \verb|vbslq_u8|(\overrightarrow{z}, \overrightarrow{\texttt{leafidx}[m]}, \overrightarrow{b})$
			\STATE $\overrightarrow{c_1} \gets \verb|vbslq_u8|(\overrightarrow{z}, \overleftarrow{m}, \overrightarrow{c_1})$
			\ENDFOR
			\STATE $\overrightarrow{c_2} \gets \verb|vrbit_u8|(\verb|vclzq_u8|(\overrightarrow{b}))$
			\STATE \textbf{return} $\verb|vmlaq_u8|(\overrightarrow{c_2}, \overrightarrow{c_1}, \overleftarrow{8})$
		\end{algorithmic}
	\end{algorithm}

	\subsection{From Ranking to Classification}
	QS and its siblings have originally been proposed to ranking ensembles. The main difference between ranking and classification ensembles is that the leaves contain not just one but multiple values.
	This needs to be addressed in the score computation. Instead of just a variable, we need an array that can store the values for every class. To add the scores of every class, we also need a loop around lines 13 to 15 of algorithm \ref{alg:qs}.
	
	Normally the scores of all classes for one instance would be stored contiguously. If we consider a block of $n$ instances for a dataset with $C$ classes and $s_{i,c}$ describes the score of class $c$ for instance $i$ , the $n \cdot C$ scores are saved as follows:
	
	{\small$s_{0,0},s_{0,1},...,s_{0,(C-1)},s_{1,0},...,s_{1,(C-1)},s_{2,0},...,s_{(n-1),(C-1)}$}
	
	This is the case for QS, VQS and RS however change the resulting data layout.
	Since these algorithms process multiple instances in parallel, the additional loop changes the order in which the scores are stored. For $n$ instances that are evaluated in parallel and $C$ classes, the scores are now stored as follows:
	
	{\small$s_{0,0},s_{1,0},...,s_{(n-1),0},s_{0,1},...,s_{(n-1),1},s_{0,2},...,s_{(n-1),(C-1)}$}

	
	\section{Quantization of trees}
	\label{sec:quant}
	As depicted in Table \ref{tab:mcus}, many MCUs do not have native floating point support. This means that floating point operations must be emulated in software rather than using the built-in hardware support leading to long execution times and higher energy consumption. There are two floating-point operations required to execute a tree ensemble: First, the individual splits of each DT in the forest are usually floating point values, and hence float comparisons $x_k \le t$ must be performed in each node in each tree. Luckily, the IEEE-754 floating-point standard guarantees that the comparisons between floating point numbers can be performed via integer operations\footnote{The comparison of two floats can be implemented as a combination of bit-by-bit comparison plus some extra handling of special cases such as \texttt{NaN} or \texttt{infinity} values.} and hence the comparisons between split values does not require hardware emulation \cite{IEEE754}. This idea has recently been explored by Hakert et al. in \cite{Hakert/etal/2022} in more detail and can be considered orthogonal to our approach here, and  both approaches can be combined with one another. Second, computing the (soft) majority vote of the ensemble requires floating point summation: Recall, that during a pre-processing step, we can scale each leaf node with potential weight $w_i$ so that the majority vote becomes a simple sum as explained in section \ref{sec:rel-work}.
	Technically, this sum is the \emph{only} arithmetic operation required to execute the entire tree ensemble. Unfortunately, there is no way to mitigate this sum, and hence floating point emulation is required if the (scaled) leaf nodes contain floats. 
	
	To solve this, we adopt a fixed-point quantization for executing decision trees. Fixed-point quantization of decision trees has already been considered in the context of FPGA accelerators \cite{Buschjaeger/Morik/2017b,Summers/etal/2020,Alcolea/Resano/2021} and for the training of DTs, e.g. by the means of histogram binning \cite{chen/etal/2016,Prokhorenkova/etal/2018,Ke/etal/2017} or through quantized boosting \cite{devos/etal/2019}. However, to the best of our knowledge, fixed-point quantization of pre-trained forests for inferencing on CPUs has not yet been studied. More specifically, we propose using fixed-point quantization for \emph{both} split nodes and the entries in the leaf values. As argued before, we may perform float comparisons using integer operations. However, a fixed-point quantization also allows us to store split thresholds in smaller words using less than $32$ bit, e.g., only $16$ bit. In this case, we can execute twice as many comparisons in VQS and RS, leading to a potential twice as fast algorithm. To this end, we propose to quantize the splits and leaf nodes during pre-processing in each tree using
	\begin{equation}
		q(x) = \lfloor s \cdot x \rfloor
	\end{equation}
	where $s \in \mathbb N_+$ is a positive scaling constant. During deployment, we only access $q(x)$ and store it in an integer variable of fixed word size. Note that the scaling constant must be chosen accordingly to the ensemble and word size. If, for example, $s = 2^{17}$ then we cannot store meaningful values in a $16$ bit variable. Similarly, consider the case in which a leaf contains a probability estimate for each class $\widehat y_i \in [0,1)$ and we use $s < M$ for a Random Forest with $w_i = \frac{1}{M}$. In this case, the scaled leaf values evaluate to
	$$
	q(x) = \left \lfloor M \cdot \frac{1}{M} \widehat y \right \rfloor = \left \lfloor \widehat y \right \rfloor = 0
	$$
	Hence, we use $s \in [M, 2^B]$ where $B$ is the maximum word that can be efficiently processed on the target hardware. 
	
	

	
	\subsection{Quantization and QuickScorer}
	By using fixed-point arithmetic for the algorithms, we double the number of instances that can be processed in parallel for VQS. The limitation for how many instances we can evaluate in parallel is the number of feature values we can compare in parallel. For the data type \texttt{float} and NEON, this is 4. By using 16 bit integers for fixed-point arithmetic we can compare 8 feature values in parallel with \verb|vcgtq_s16|.
	The elements of the mask that this comparison returns are only 16 bits wide. But for VQS the masks need to have the size of the number of leaves $L$. Since VQS is applied on ensembles with $L = 32$ and $L = 64$ in our experiments, we need to extend the masks to these numbers. This is done with a combination of \verb|vget_low_s16|/\verb|vget_high_s16| and \verb|vmovl_s16|. The first intrinsic returns the lower or higher half of a SIMD register, and the latter then extends the mask correctly. This returns two masks with 32 bit elements. For $L = 64$ the two masks then must again be extended to four masks with 64 bit elements with the intrinsics \verb|vget_low_s32|/\verb|vget_high_s32| and \verb|vmovl_s32|.
	For the score computation, the parallelism of adding the scores is also doubled since we can add eight 16 bit values at once.
	
	For the {\rapidscorer} algorithm, the number of instances being processed in parallel is not limited by the number of values that can be compared in parallel but by the number of bytes that fit into one SIMD register. For NEON, these are 16 instances. Therefore, {\rapidscorer} needs to use four compare operations, and then it combines the resulting bitmasks. By using 16 bit fixed-point values, the number of comparisons is reduced to two, and the resulting two masks must only be combined once.
	The score computation for {\rapidscorer} also profits from the increased parallelism in adding the scores. The previously explained vectorized search for the correct exit leaf does not profit from this change since it is not dependent on the representation of feature or score values.

	\section{Experiments}
	\label{sec:experiments}
	In our experimental evaluation, we study the behavior of {\quickscorer} and its siblings on various datasets and different hardware configurations. We are specifically interested in 
	\begin{itemize}
		\item \textbf{Question 1:} Do {\quickscorer}, {\vquickscorer} and {\rapidscorer} perform as well on different ARM devices for ranking as they do on Intel?
		\item \textbf{Question 2:} What is the impact of quantization on the accuracy of tree ensembles for classification problems?
		\item \textbf{Question 3:} What is the performance of (quantized) {\quickscorer}, {\vquickscorer} and {\rapidscorer} on ARM devices for classification?
	\end{itemize}
	
	For our first experiment in the context of ranking, we follow the experimental setup in \cite{lucchese/etal/2015}. We train Gradient Boosted trees with $\{1000,5000,10000,20000\}$ tress each with at most $\{32,64\}$ leaves on the MSN dataset \cite{MSN} using XGBoost \cite{chen/etal/2016}. For the second and third experiments in the context of classification, we train a Random Forest with $1024$ trees in which each tree has at most $\{32, 64\}$ leaves using scikit-learn \cite{pedregosa2011scikit}. 
	
	
	The classification experiments are performed on the publicly available datasets Magic \cite{Magic} with 10 features and 2 classes, Adult \cite{Adult} with 108 features and 2 classes, EEG \cite{EEG} with 14 features and 2 classes, MNIST \cite{MNIST} with 784 features and 10 classes and Fashion \cite{Fashion} with 784 features and 10 classes. In all experiments, we use an $80/20$ train/test, except the dataset comes with a pre-computed train/test split (e.g., MNIST and Fashion), which we use in that case. All experiments are executed on a Raspberry Pi 3 Model B+ and an Odroid-XU4. The Raspberry Pi has a Broadcom BCM2837B0 (ARM Cortex-A53) processor. It clocks at 1.4 GHz and has 1 GiB of RAM and a shared L2 cache with 512 KB. The Odroid uses a Samsung Exynos 5422 processor with a Big. Little architecture combining an ARM Cortex A15 clocked at 2GHz, and an ARM Cortex A7 clocked at 1.4GHz. It has 2 GB of RAM. As a baseline we use the previously mentioned \textsc{if-else} implementation and \textsc{native} implementation (sometimes also called \textsc{PRED}, see e.g. \cite{Asadi/etal/2014}) using FastInference \cite{FastInference}. FastInference is a code generator that generates model-specific \texttt{C++} code that is then statically compiled for the execution on each ARM device using a cross-compiler. We compare this against our own implementations for QS, VQS, and RS. During all experiments, we made sure all implementations produced the \emph{same} prediction for the same ensemble.
	
	\subsection{Q1: What is the performance of QS/VQS/RS on ARM for ranking problems?}
	In the first experiment, we study the performance of QS and its siblings on ARM for ranking tasks. This experiment closely follows the experimental setup from \cite{lucchese/etal/2015} but is performed on ARM devices instead of an Intel machine.
	Table \ref{table:ranking:benchmark:neon} shows the average runtime per instance in $\mu s$ for {\quickscorer} (QS), {\vquickscorer} (VQS), {\rapidscorer} (RS), If-Else (IE), Native (NA) on the Raspberry Pi (ARM Cortex A53, Top group) and Odroid-XU4 (Samsung Exynos 5422, Bottom group) on the MSN dataset. Orange marks the best implementation (smaller is better). The speed-up compared to the Native (NA) implementation is given in parenthesis. Looking at the performance of the Raspberry Pi (top group), one can see that RS is clearly the best implementation. It offers the largest speed-up ranging from $1.6$ to $5.8$ on all experiments, followed by VQS, which offers a speed-up of around $0.8 - 2.5$. Somewhat surprisingly, the runtime does not increase linearly with the number of trees. For example, QS has a runtime of $173 \mu s$ for $1000$ trees which increases by roughly $10$ times to $1562 \mu s$ for $5000$ trees and then further increases roughly $3$ times to $4522 \mu s$ for $10~000$ trees. It is also noteworthy that the doubling of the number of leaf nodes from $32$ to $64$ does not lead to the doubling in runtime but only leads to a slight increase in the runtime. Looking at the performance on the Odroid-XU4 (bottom group), one can see a more fragmented behavior. Here, RS dominates for $64$ leaves, but VQS is now sometimes the best method for $32$ leaves. RS and VQS offer a speed-up of around $1.4 - 9.4$ in some instances, which is a much larger speed-up compared to the Raspberry Pi.
	
	There seem to be some architectural differences between the Cortex A53 and the Exynos 5422 that impact the performance of the implementations.
	We conclude that, for the best performance, the combination between forest, device \emph{and} implementation is important.

	
	\begin{table*}[h!]
		\caption{Runtime per instance in $\mu s$ for {\quickscorer} (QS), {\vquickscorer} (VQS), {\rapidscorer} (RS), If-Else (IE), Native (NA) on the Raspberry Pi (ARM Cortex A53, Top group) and Odroid-XU4 (Samsung Exynos 5422, Bottom group) on the MSN dataset. Orange marks the best implementation (smaller is better). The speed-up compared to the Native (NA) implementation is given in parenthesis. Best viewed in color.}
		\label{table:ranking:benchmark:neon}
		\small
		\centering
		\begin{tabular}{ p{1.9cm} | p{1.3cm} | p{0.4cm} | p{2cm} | p{2cm} | p{2cm} | p{2cm} }
			\hline
			\multirow{2}{1.9cm}{Processor} & \multirow{2}{1.3cm}{Algorithm} & \multirow{2}{0.4cm}{$L$} & \multicolumn{4}{c}{Number of trees}\\
			\cline{4-7}
			&&& 1000 & 5000 & 10000 & 20000 \\
			\hline\hline
			\multirow{10}{1.9cm}{ARM Cortex A53} & RS & \multirow{5}{0.4cm}{32} & \textcolor{orange}{117.9 (2.6x)} & \textcolor{orange}{511.7 (5.8x)} & \textcolor{orange}{1542.6 (4.0x)} & \textcolor{orange}{4025.9 (3.0x)} \\
			& VQS & & 120.6 (2.5x) & 1185.1 (2.5x) & 2489.7 (2.5x) & 5510.0 (2.2x) \\
			& QS & & 173.4 (1.8x) & 1562.8 (1.9x) & 4522.1 (1.4x) & 10601.7 (1.2x) \\
			& IE & & 430.8 (0.7x) & 3649.6 (0.8x) & 7309.0 (0.8x) & 14456.7 (0.8x) \\
			& NA & & 306.6 (-) & 2993.1 (-) & 6140.6 (-) & 12264.0 (-) \\
			\cline{2-7}
			& RS & \multirow{5}{0.4cm}{64} & \textcolor{orange}{219.9 (2.9x)} & \textcolor{orange}{1557.3 (2.5x)} & \textcolor{orange}{4220.1 (1.9x)} & \textcolor{orange}{9985.9 (1.6x)} \\
			& VQS & & 372.9 (1.7x) & 2709.1 (1.4x) & 5705.6 (1.4x) & 19988.4 (0.8x) \\
			& QS & & 373.3 (1.7x) & 4086.1 (1.0x) & 8695.3 (0.9x) & 19339.9 (0.8x) \\
			& IE & & 644.1 (1.0x) & 4267.7 (0.9x) & 8572.5 (0.9x) & 17354.4 (0.9x) \\
			& NA & & 626.8 (-) & 3919.6 (-) & 8000.6 (-) & 16126.7 (-) \\
			\hline\hline
			\multirow{10}{1.9cm}{Samsung Exynos 5422} & RS & \multirow{5}{0.4cm}{32} & \textcolor{orange}{30.9 (3.5x)} & 265.8 (6.4x) & 622.7 (9.1x) & \textcolor{orange}{1316.0 (9.3x)} \\
			& VQS & & 35.7 (3.0x) & \textcolor{orange}{204.1 (8.3x)} & \textcolor{orange}{600.5 (9.4x)} & 1461.4 (8.4x) \\
			& QS & & 57.9 (1.9x) & 331.1 (5.1x) & 1027.4 (5.5x) & 2087.7 (5.8x) \\
			& IE & & 92.8 (1.2x) & 1675.9 (1.0x) & 3329.4 (1.7x) & 7202.9 (1.7x) \\
			& NA & & 108.8 (-) & 1699.3 (-) & 5650.2 (-) & 12204.3 (-) \\
			\cline{2-7}
			& RS & \multirow{5}{0.4cm}{64} & \textcolor{orange}{69.8 (1.8x)} & \textcolor{orange}{625.8 (4.1x)} & \textcolor{orange}{1332.5 (5.4x)} & \textcolor{orange}{3037.9 (4.9x)} \\
			& VQS & & 92.9 (1.4x) & 722.4 (3.5x) & 1807.5 (4.0x) & 6052.7 (2.5x) \\
			& QS & & 134.1 (0.9x) & 1160.7 (2.2x) & 2143.9 (3.3x) & 5217.3 (2.9x) \\
			& IE & & 154.6 (0.8x) & 2185.3 (1.2x) & 4305.0 (1.7x) & 9808.7 (1.5x) \\
			& NA & & 126.3 (-) & 2535.1 (-) & 7152.1 (-) & 14980.4 (-) \\
			\hline
		\end{tabular}
	\end{table*}

	\subsection{Q2: What is the impact of quantization on tree ensembles?}
	In the second experiment, we compare the classification accuracies of quantized RF to `regular' RF using floating points.
	We also investigate the impact of quantization on the merging of equivalent nodes in {\rapidscorer}.
	In this experiment we use 16 bit integers ( \texttt{short} variables) and 32 bit floats (\texttt{float} variables) to store leaf nodes and splits. For quantization, we scale each float with $s = 2^{15}$ and round it towards the next smallest integer as discussed in section \ref{sec:quant}. For space reasons, we focus on RFs with $1024$ trees and $64$ leaf nodes. Additional results for RFs with $32$ leaf nodes can be found in the appendix. Table \ref{table:classification:accuracies} shows the results for this experiment. As one can see, quantization does not seem to have a large impact on the accuracy, except on the EEG dataset, where the quantization of the leaf values leads to a drop of nearly 4 percentage points. 
	Table \ref{table:nodesmerging} shows the effects of this quantization on the node merging of {\rapidscorer} that also helps to explain the accuracy drop on the EEG dataset. One can see that there is a drastic change in the number of unique nodes in the ensemble on the EEG dataset when comparing float and fixed points. The number of unique nodes in all datasets remains close when changing from floating point splits to fixed point splits except on the EEG dataset. Here, the number of unique nodes drops by nearly $50$ \% when going from floats to fixed point data types partly explaining the drop in accuracy in Table \ref{table:classification:accuracies}. 
	\begin{table}[h]
		\setlength{\tabcolsep}{8pt}
		\centering
		\caption{Accuracy of different quantizations for an RF with $1024$ trees and $64$ leaf nodes. Int16 denotes 16 bit integer and float denotes a 32 bit float.}
		\label{table:classification:accuracies}
		\resizebox{\linewidth}{!}{
			\begin{tabular}{@{}rrrrr@{}}
				\toprule
				Dataset & \begin{tabular}[c]{@{}r@{}}split: float\\ leaf: float\end{tabular} & \begin{tabular}[c]{@{}r@{}}split: float\\ leaf: int16\end{tabular} & \begin{tabular}[c]{@{}r@{}}split: int16\\ leaf: float\end{tabular} & \begin{tabular}[c]{@{}r@{}}split: int16\\ leaf: int16\end{tabular} \\ \midrule
				Adult & 84.66\% & 84.66\% & 84.66\% & 84.66\% \\
				EEG & 78.37\% & 78.37\% & 74.27\% & 74.27\% \\
				Fashion & 79.87\% & 79.86\% & 79.87\% & 79.86\% \\
				Magic & 84.60\% & 84.57\% & 84.57\% & 84.54\% \\
				MNIST & 89.24\% & 89.21\% & 89.24\% & 89.21\% \\ \bottomrule
			\end{tabular}
		}
	\end{table}
	
	\begin{table}[h]
		\setlength{\tabcolsep}{8pt}
		\centering
		\caption{{Percentage of unique nodes that are kept after merging of equivalent nodes on the classification datasets.}}
		\label{table:nodesmerging}
		\footnotesize
		\begin{tabular}{@{}llrrrr@{}}
			\toprule
			\multirow{2}{*}{Dataset} & \multirow{2}{*}{Type} & \multicolumn{4}{c}{Number of Trees}   \\ \cmidrule(l){3-6} 
			&                       & 128     & 256     & 512     & 1024    \\ \midrule
			\multirow{2}{*}{Adult}   & float                 & 12.1 \% & 9.5 \%  & 7.6\%   & 6.0 \%   \\
			& quant                 & 12.3 \% & 9.6 \%  & 7.6 \%  & 5.9 \%\vspace{0.1cm} \\
			\multirow{2}{*}{EEG}     & float                 & 52.2 \% & 39.5 \% & 28.5 \% & 19.4 \% \\
			& quant                 & 28.6 \% & 19.8 \% & 13.3 \% & 8.4 \%\vspace{0.1cm} \\
			\multirow{2}{*}{Fashion} & float                 & 84.6 \% & 76.3 \% & 66.5 \% & 55.5\%  \\
			& quant                 & 84.6 \% & 76.3 \% & 66.5 \% & 55.6 \%\vspace{0.1cm} \\
			\multirow{2}{*}{Magic}   & float                 & 88.7 \% & 81.6 \% & 73.5 \% & 63.5 \% \\
			& quant                 & 86.7 \% & 78.7 \% & 69.3 \% & 58.3 \%\vspace{0.1cm} \\
			\multirow{2}{*}{MNIST}   & float                 & 65.6 \% & 56.6 \% & 47.7 \% & 39.1 \% \\
			& quant                 & 65.6 \% & 56.6 \% & 47.7 \% & 39.1 \% \\ \bottomrule
		\end{tabular}
		
	\end{table}
	
	
	\subsection{Q3: What is the performance of (quantized) QS/VQS/RS on ARM for classification problems?}
	
	In the third experiment, we study the inference speed of (quantized) \textsc{Quick-Scorer} and its siblings for classification tasks on different ARM devices. Similar to before, we train RF models with $1024$ trees using $\{32, 64\}$ leaf nodes on the \{Adult, EEG, Fashion, Magic, MNIST\} dataset and execute the ensembles with (quantized) \{QS, VQS, RS, NA, IE\} implementations. For quantization, we use 16 bit integers by scaling each float with $s = 2^{15}$ and then by rounding  towards the next smallest integer. Since quantization did not have any negative impact on the accuracy except on the EEG data, we  quantize both splits and leaf values in this experiment. We add the prefix `q' to each implementation to denote its quantized variation. We first study the average speed-up of each method against a baseline implementation, namely the native (NA) implementation using float variables. 
	
	\begin{table*}[h]
		\caption{Runtime per instance in $\mu s$ for {\quickscorer} (QS), {\vquickscorer} (VQS), {\rapidscorer} (RS), If-Else (IE), Native (NA) on the Raspberry Pi (ARM Cortex A53, top group) and Odroid-XU4 (Samsung Exynos 5422, bottom group) on various classification datasets. Orange marks the best implementation (smaller is better). The speed-up compared to the Native (NA) implementation is given in parenthesis. The ensembles consist of 1024 trees with 64 leaves. Best viewed in color.}
		\label{table:classification:pi}
		\centering
		\small
		\begin{tabular}{ p{1.9cm} | p{1.3cm} | p{1.9cm} | p{1.9cm} | p{1.9cm} | p{1.9cm} | p{1.9cm} }
			\hline
			\multirow{2}{1.9cm}{Processor} & \multirow{2}{1.3cm}{Algorithm} & \multicolumn{5}{c}{Dataset}\\
			\cline{3-7}
			&& Magic & MNIST & Adult & EEG & Fashion \\
			\hline\hline
			\multirow{10}{1.9cm}{ARM Cortex A53} & RS & 381.9 (2.0x) & \textcolor{orange}{465.1 (2.5x)} & 186.2 (3.7x) & 242.1 (2.9x) & \textcolor{orange}{508.2 (2.3x)} \\
			& VQS & 519.8 (1.5x) & 858.1 (1.4x) & 308.4 (2.2x) & 454.1 (1.5x) & 842.9 (1.4x) \\
			& QS & 513.4 (1.5x) & 743.6 (1.6x) & 246.1 (2.8x) & 406.4 (1.7x) & 871.6 (1.4x) \\
			& IE & 1036.7 (0.7x) & 1377.1 (0.9x) & 575.2 (1.2x) & 931.5 (0.7x) & 1327.3 (0.9x) \\
			& NA & 770.2 (-) & 1176.1 (-) & 689.5 (-) & 693.0 (-) & 1181.9 (-) \\
			\cline{2-7}
			& qRS & \textcolor{orange}{335.0 (2.3x)} & 475.6 (2.5x) & \textcolor{orange}{173.0 (4.0x)} & \textcolor{orange}{191.8 (3.6x)} & 542.2 (2.2x) \\
			& qVQS & 410.2 (1.9x) & 735.1 (1.6x) & 264.3 (2.6x) & 411.7 (1.7x) & 652.2 (1.8x) \\
			& qQS & 426.7 (1.8x) & 592.6 (2.0x) & 201.9 (3.4x) & 316.3 (2.2x) & 718.0 (1.6x) \\
			& qIE & 480.4 (1.6x) & 847.4 (1.4x) & 579.7 (1.2x) & 445.2 (1.6x) & 895.6 (1.3x) \\
			& qNA & 417.2 (1.8x) & 779.2 (1.5x) & 370.0 (1.9x) & 377.7 (1.8x) & 794.3 (1.5x) \\
			
			\hline\hline
			
			\multirow{10}{1.9cm}{Samsung Exynos 5422} & RS & 229.7 (0.6x) & 299.5 (1.3x) & 54.1 (2.6x) & 89.0 (1.7x) & 322.8 (1.1x) \\
			& VQS & 120.2 (1.2x) & 266.3 (1.5x) & 94.2 (1.5x) & 103.0 (1.5x) & 255.2 (1.4x) \\
			& QS & 185.3 (0.8x) & 309.8 (1.3x) & 97.4 (1.5x) & 160.0 (1.0x) & 340.6 (1.1x) \\
			& IE & 300.6 (0.5x) & 818.1 (0.5x) & 140.0 (1.0x) & 291.7 (0.5x) & 890.4 (0.4x) \\
			& NA & 146.7 (-) & 387.7 (-) & 141.4 (-) & 153.6 (-) & 366.4 (-) \\
			\cline{2-7}
			& qRS & 204.2 (0.7x) & 294.6 (1.3x) & \textcolor{orange}{46.1 (3.1x)} & \textcolor{orange}{54.3 (2.8x)} & 329.3 (1.1x) \\
			& qVQS & 104.3 (1.4x) & 175.0 (2.2x) & 72.3 (2.0x) & 134.3 (1.1x) & 176.4 (2.1x) \\
			& qQS & 166.4 (0.9x) & 223.4 (1.7x) & 84.4 (1.7x) & 128.5 (1.2x) & 274.6 (1.3x) \\
			& qIE & \textcolor{orange}{98.6 (1.5x)} & 417.7 (0.9x) & 102.2 (1.4x) & 98.3 (1.6x) & 391.9 (0.9x) \\
			& qNA & 124.4 (1.2x) & \textcolor{orange}{165.3 (2.3x)} & 119.6 (1.2x) & 126.0 (1.2x) & \textcolor{orange}{170.4 (2.2x)} \\
			\hline
		\end{tabular}
	\end{table*}
	
	We start our analysis by looking at the detailed results for $1024$ trees and $64$ leaves. Table \ref{table:classification:pi} shows the runtime per instance in $\mu s$ of {\quickscorer} (QS), {\vquickscorer} (VQS), {\rapidscorer} (RS), If-Else (IE), Native (NA) on the Raspberry Pi (ARM Cortex A53, Top group) and Odroid-XU4 (Samsung Exynos 5422, Bottom group) on various classification datasets.
	For the Raspberry Pi (top group) RS is the best implementation with speedups of $2.0 - 3.7$ and its quantized variant with speedups of $2.2 - 4.0$. Surprisingly, QS seems to outperform VQS on the Raspberry Pi with speedups of $1.4 - 2.8$. The quantization of the ensemble seems to make IE and NA much faster. IE performs similarly to NA with speedups of $0.7 - 1.2$ while qIE has speedups of $1.2 - 1.6$. qNA also performs better with speedups of ranging from $1.5$ to $1.9$ over NA.
	Looking at the results of the Odroid (bottom group), one can see that there is no clear winner here. VQS is faster than RS on all but the adult dataset for the float variants making it the fastest non-quantized implementation. Looking at the quantized version, the picture becomes more fragmented. The quantized versions of IE and NA seem to be gaining the most from quantization. While IE has a speedup of $0.5 - 1.0$, qIE is $0.9$ to $1.6$ times faster. qNA has a speedup of $1.2 - 2.3$ making them the fastest method on two datasets MNIST and Fashion. Interestingly, q(V)QS is now also outperformed by qRS on the adult and EEG datasets making q(V)QS the overall slowest method. 
	
\begin{figure*}[h]
		\centering
		\begin{subfigure}{1.0\columnwidth}
			\centering
			\includegraphics[width=7.5cm, height=5.8cm]{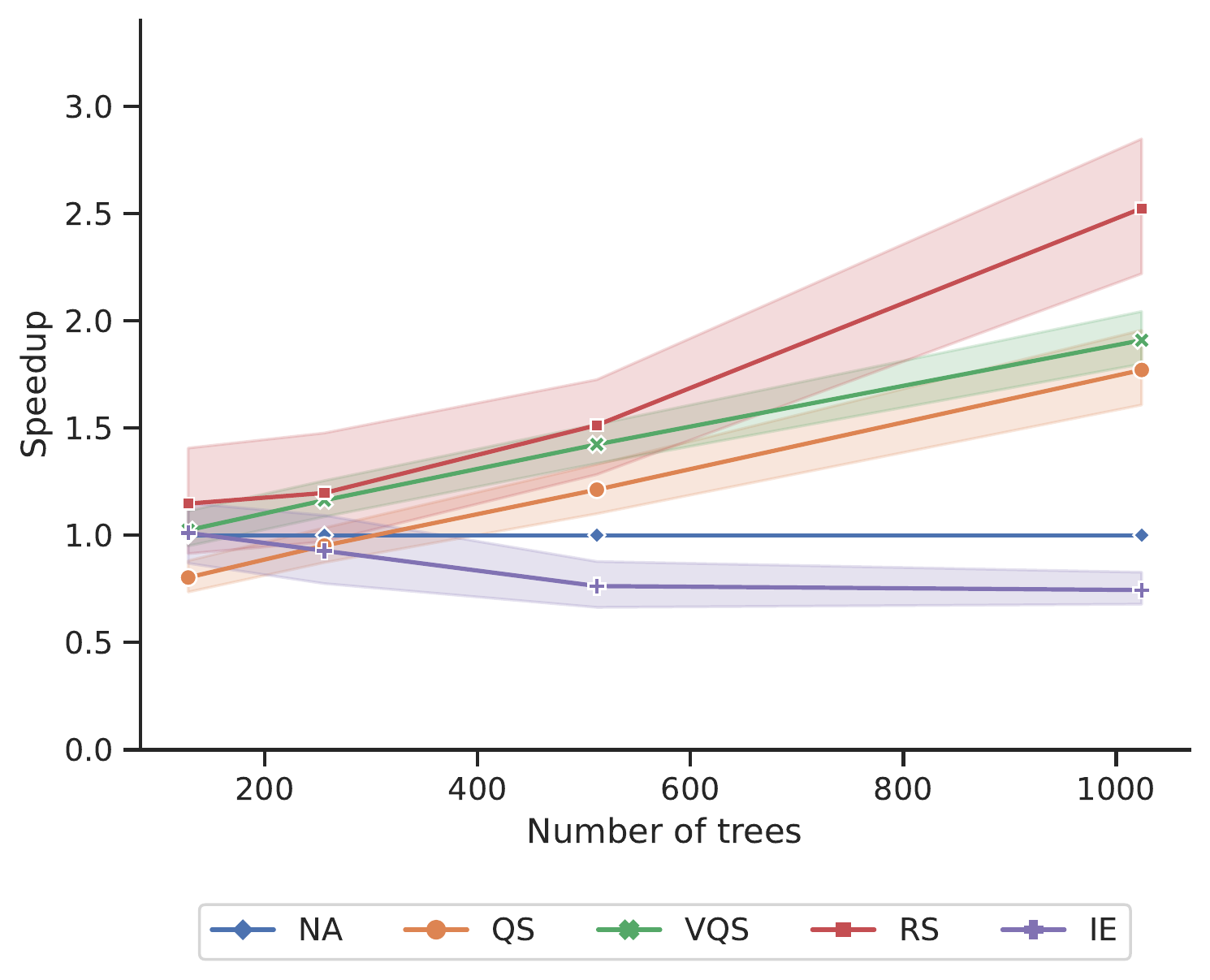}
			\label{fig:classification:results:float}
		\end{subfigure}
		\begin{subfigure}{1.0\columnwidth}
			\centering
			\includegraphics[width=8cm, height=5.8cm]{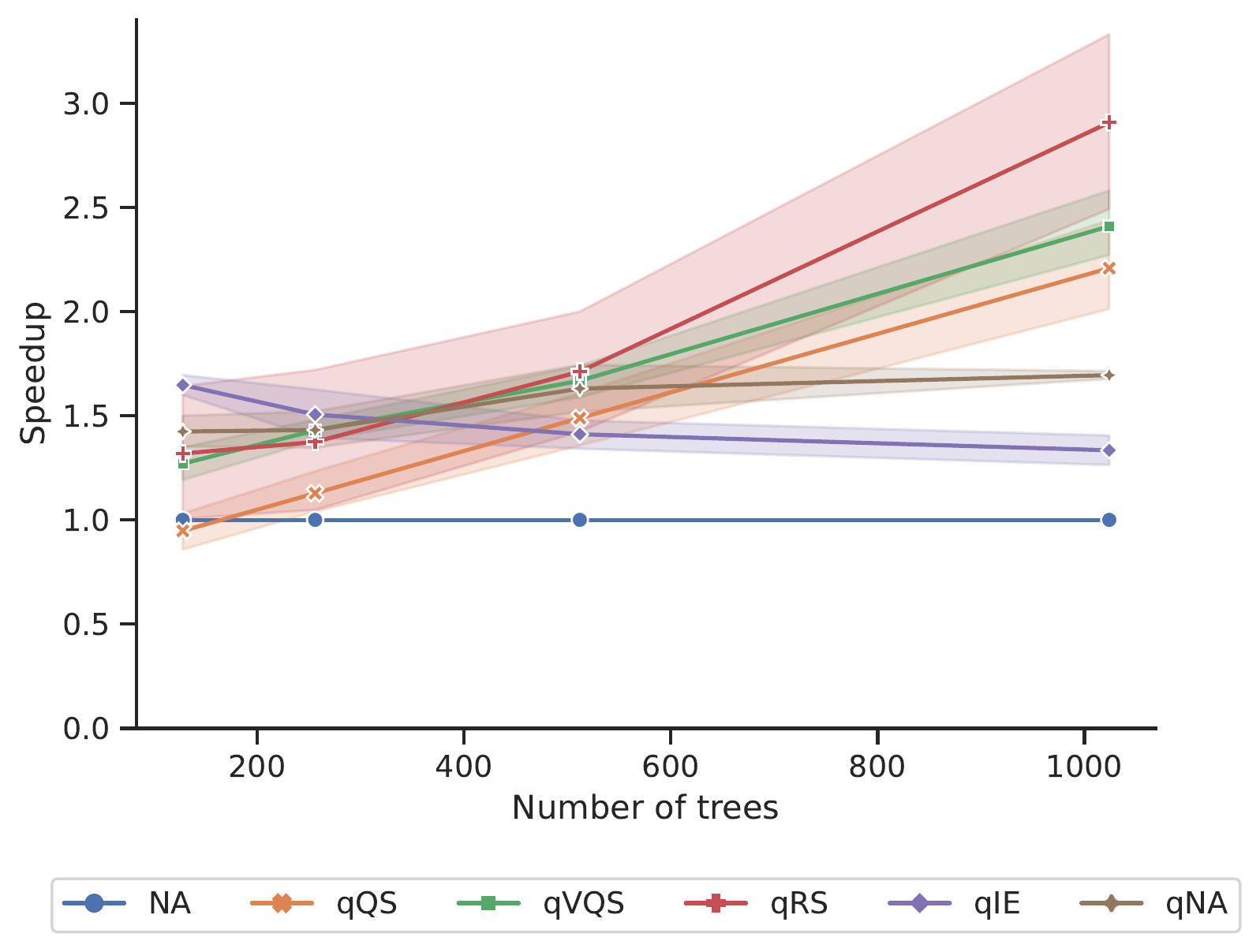}
			
			
			\label{fig:classification:results:quantized}
		\end{subfigure}
		\caption{Average speed-up of the different implementations (top) over a native (NA) implementation and average speed-up of the different quantized implementations (bottom) over a native (NA) implementation across a different number of trees. Results are averaged across the 5 classification datasets and two ARM devices. Larger is better. The error band shows the standard deviation.}
		\label{fig:classification:results}
	\end{figure*}

	To give a more complete picture we investigate the impact of the number of trees for the different implementations.
	Figure \ref{fig:classification:results} shows the average speed-up of the different float implementations (left side) and the average speed-up of the different quantized implementations (right side) for different numbers of trees. The average was taken wrt. to the number of leaves and across the different datasets. Note that in both cases, we compare against the float native implementation hence NA is depicted as a constant line. Second, we made sure that the scaling on the y-axis is the same so that both plots can be compared in terms of speed-up. As one can see, RS and qRS show the most dramatic speed-up ranging from just above 1 to almost $2.5$. Similar, (q)QS and (q)VQS also show consistent speed-ups although less dramatic than (q)RS. The (q)IE and (q)NA implementations have more mixed behavior. Vanilla IE is slower than NA, but using quantization leads to a speed-up of around $1.5$. Similarly, qNA also achieves a consistent speed-up of around $1.5$ for more than $400$ trees. We conclude that quantization leads to consistent speed-up across all configurations. IE and NA seem to benefit the most from quantization, whereas QS and its siblings receive a smaller performance boost from using quantized models. 
	
\begin{figure}[h]
		\centering
		\begin{subfigure}{1.0\columnwidth}
			\centering
			\includegraphics[width=0.8\columnwidth, keepaspectratio]{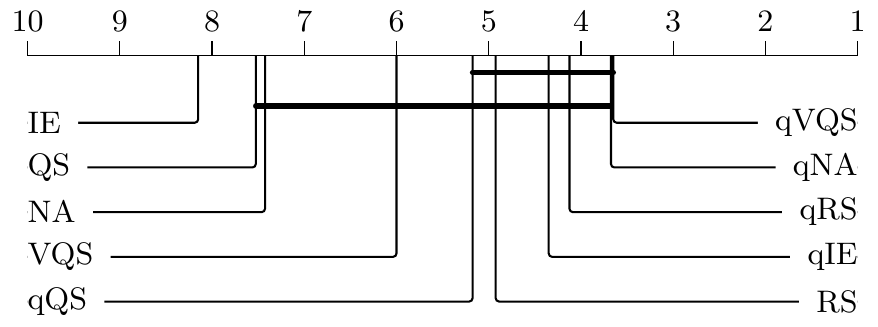}
			\label{fig:classification:cdd2:pi}
		\end{subfigure}
		
		\begin{subfigure}{1.0\columnwidth}
			\centering
			\includegraphics[width=0.8\columnwidth, keepaspectratio]{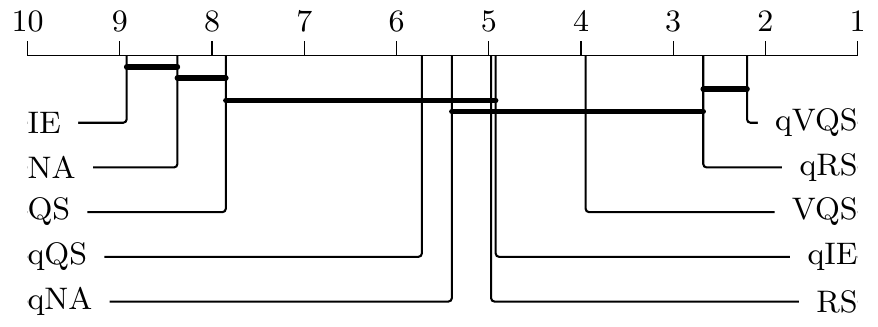}
			\label{fig:classification:cdd2:odroid}
		\end{subfigure}%
		\caption{Critical Difference Diagram for the average inferencing speed. The top plot depicts the results for the Raspberry Pi and the bottom plot shows the results for the Odroid-XU4. For all statistical tests, $p=0.95$ has been used.}
		\label{fig:classification:cdd2}
	\end{figure}

	To give a more statistical interpretation to these results we now plot the average runtime per observation of each implementation on each dataset in a Critical Difference Diagram (CD Diagram). CD diagrams have been proposed as a method to compare multiple methods across multiple datasets \cite{demvsar/2006}. A CD diagram ranks each method according to its performance (here inferencing speed) on each dataset. Then, a Friedman-Test is performed to determine if there is a statistical difference between the average rank of each method. If this is the case, a subsequent pairwise Wilcoxon-Test between all methods is used to check whether there is a statistical difference between the two methods (see \cite{Benavoli/etal/2016}). CD diagrams visualize this evaluation by plotting the average rank of each method on the x-axis and connecting all methods whose performances are statistically \emph{not} distinguishable from each other via a vertical bar. Differently put, implementations that belong to the same connected clique have a statistically similar inferencing speed. Figure \ref{fig:classification:cdd2} shows the resulting CD Diagram. The top plot shows the rank of each implementation on the Odroid-XU4 and the bottom plot depicts the ranks for the Raspberry Pi. As indicated by the previous experiments, quantization generally leads to faster implementations. Similar, (q)VQS and (q)RS generally seem to be the best choice for the Odroid-XU4, where qVQS and qRS are the fastest methods. Interestingly, VQS (rank 3) and RS (rank 5) also outperform some of the quantized versions of the other algorithms such as qQS and qNA. Looking at the results for the Raspberry Pi we see a slightly different picture. Here, qVQS and qNA are extremely close to each other nearly ranking in the same place. Moreover, placings are generally closer compared to the Odroid-XU4 where there is a more even spread of ranks.
	
	This experiment further supports our hypothesis from the first experiment: For each combination of hardware platform as well as dataset and forest, there seems to be a unique implementation best suited for inferencing.  
	
	\section{Conclusion}
	\label{sec:conclusion}
	Model deployment and the continuous execution of ML models becomes more and more an important issue, especially in the context of the Internet Of Things. Tree ensembles such as Random Forests or Gradient Boosted Trees are ideal candidates for IoT applications due to their excellent performance and small resource consumption. However, despite these advantages, a careful implementation of tree ensembles must be provided to deploy them onto embedded IoT devices. In this paper we revisited the {\quickscorer} algorithm family and presented the first adaption of {\quickscorer}, {\vquickscorer} and {\rapidscorer} for ARM CPUs. We showed that with careful implementation many of the performance benefits of QS, VQS, and RS can also be utilized on ARM CPUs.
	Moreover, we studied the quantization of tree ensembles to further reduce the memory consumption of individual trees while having a minor impact on the classification accuracy. Last, we investigated the performance of our implementation on different ARM devices, namely a Raspberry Pi and an Odroid-XU4. Our study shows that the best implementation depends on both, the device as well as the specific tree ensemble, highlighting that subtle architectural differences can play a major role in the deployment of ML models.

	\bibliography{literature}

\begin{thebibliography}{40}
\providecommand{\natexlab}[1]{#1}
\providecommand{\url}[1]{\texttt{#1}}
\expandafter\ifx\csname urlstyle\endcsname\relax
  \providecommand{\doi}[1]{doi: #1}\else
  \providecommand{\doi}{doi: \begingroup \urlstyle{rm}\Url}\fi

\bibitem[Adu()]{Adult}
Adult dataset.
\newblock \url{https://archive.ics.uci.edu/ml/datasets/adult}.

\bibitem[EEG()]{EEG}
Eeg dataset.
\newblock \url{https://archive.ics.uci.edu/ml/datasets/eeg+database}.

\bibitem[Fas()]{Fashion}
Fashion mnist dataset.
\newblock \url{https://github.com/zalandoresearch/fashion-mnist}.

\bibitem[MNI()]{MNIST}
Mnist dataset.
\newblock \url{http://yann.lecun.com/exdb/mnist/}.

\bibitem[MSN()]{MSN}
Microsoft learning to rank.
\newblock \url{https://tinyurl.com/2p96nubu}.

\bibitem[Mag()]{Magic}
Magic dataset.
\newblock \url{https://archive.ics.uci.edu/ml/datasets/magic+gamma+telescope}.

\bibitem[IEE(2019)]{IEEE754}
Ieee standard for floating-point arithmetic.
\newblock \emph{IEEE Std 754-2019 (Revision of IEEE 754-2008)}, pp.\  1--84,
  2019.
\newblock \doi{10.1109/IEEESTD.2019.8766229}.

\bibitem[Alcolea \& Resano(2021)Alcolea and Resano]{Alcolea/Resano/2021}
Alcolea, A. and Resano, J.
\newblock Fpga accelerator for gradient boosting decision trees.
\newblock \emph{Electronics}, 10\penalty0 (3):\penalty0 314, 2021.

\bibitem[Asadi et~al.(2014)Asadi, Lin, and de~Vries]{Asadi/etal/2014}
Asadi, N., Lin, J., and de~Vries, A.~P.
\newblock Runtime optimizations for tree-based machine learning models.
\newblock \emph{IEEE Transactions on Knowledge and Data Engineering},
  26\penalty0 (9):\penalty0 2281--2292, Sept 2014.
\newblock ISSN 1041-4347.
\newblock \doi{10.1109/TKDE.2013.73}.

\bibitem[Barros et~al.(2015)Barros, de~Carvalho, and Freitas]{Barros/etal/2015}
Barros, R.~C., de~Carvalho, A. C. P. L.~F., and Freitas, A.~A.
\newblock \emph{Decision-Tree Induction}, pp.\  7--45.
\newblock Springer International Publishing, Cham, 2015.
\newblock ISBN 978-3-319-14231-9.

\bibitem[Benavoli et~al.(2016)Benavoli, Corani, and
  Mangili]{Benavoli/etal/2016}
Benavoli, A., Corani, G., and Mangili, F.
\newblock Should we really use post-hoc tests based on mean-ranks?
\newblock 17:\penalty0 5:1--5:10, 2016.
\newblock URL \url{http://jmlr.org/papers/v17/benavoli16a.html}.

\bibitem[Branco et~al.(2019)Branco, Ferreira, and Cabral]{branco/etal/2019}
Branco, S., Ferreira, A.~G., and Cabral, J.
\newblock Machine learning in resource-scarce embedded systems, fpgas, and
  end-devices: A survey.
\newblock \emph{Electronics}, 8\penalty0 (11):\penalty0 1289, 2019.

\bibitem[Buschj\"{a}ger \& Morik(2018)Buschj\"{a}ger and
  Morik]{Buschjaeger/Morik/2017b}
Buschj\"{a}ger, S. and Morik, K.
\newblock Decision tree and random forest implementations for fast filtering of
  sensor data.
\newblock \emph{IEEE Transactions on Circuits and Systems I: Regular Papers},
  65-I\penalty0 (1):\penalty0 209--222, January 2018.
\newblock URL \url{https://doi.org/10.1109/TCSI.2017.2710627}.

\bibitem[Buschjäger()]{FastInference}
Buschjäger, S.
\newblock Fastinference.
\newblock URL \url{https://github.com/sbuschjaeger/fastinference/}.

\bibitem[Buschjäger \& Morik(2021)Buschjäger and
  Morik]{Buschjaeger/Morik/2021b}
Buschjäger, S. and Morik, K.
\newblock Improving the accuracy-memory trade-off of random forests via
  leaf-refinement, 2021.
\newblock URL \url{https://arxiv.org/abs/2110.10075}.

\bibitem[Buschjäger et~al.(2018)Buschjäger, Chen, Chen, and
  Morik]{Buschjaeger/2018a}
Buschjäger, S., Chen, K.-H., Chen, J.-J., and Morik, K.
\newblock Realization of random forest for real-time evaluation through tree
  framing.
\newblock In \emph{The IEEE International Conference on Data Mining series
  (ICDM)}, November 2018.

\bibitem[Chen et~al.(2022)Chen, Hsu, Hakert, Buschj\"ager, Lee, Lee, Morik, and
  Chen]{Chen/etal/2022}
Chen, K.-H., Hsu, C.-H., Hakert, C., Buschj\"ager, S., Lee, C.-L., Lee, J.-K.,
  Morik, K., and Chen, J.-J.
\newblock Efficient realization of decision trees for real-time inference.
\newblock \emph{ACM Transactions on Embedded Computing Systems}, 2022.

\bibitem[Chen \& Guestrin(2016)Chen and Guestrin]{chen/etal/2016}
Chen, T. and Guestrin, C.
\newblock Xgboost: {A} scalable tree boosting system.
\newblock In Krishnapuram, B., Shah, M., Smola, A.~J., Aggarwal, C.~C., Shen,
  D., and Rastogi, R. (eds.), \emph{Proceedings of the 22nd {ACM} {SIGKDD}
  International Conference on Knowledge Discovery and Data Mining, San
  Francisco, CA, USA, August 13-17, 2016}, pp.\  785--794. {ACM}, 2016.
\newblock \doi{10.1145/2939672.2939785}.
\newblock URL \url{https://doi.org/10.1145/2939672.2939785}.

\bibitem[Choudhary et~al.(2020)Choudhary, Mishra, Goswami, and
  Sarangapani]{choudhary/etal/2020}
Choudhary, T., Mishra, V., Goswami, A., and Sarangapani, J.
\newblock A comprehensive survey on model compression and acceleration.
\newblock \emph{Artificial Intelligence Review}, 2020.

\bibitem[Dem{\v{s}}ar(2006)]{demvsar/2006}
Dem{\v{s}}ar, J.
\newblock Statistical comparisons of classifiers over multiple data sets.
\newblock \emph{The Journal of Machine Learning Research}, 7:\penalty0 1--30,
  2006.

\bibitem[Devos et~al.(2019)Devos, Meert, and Davis]{devos/etal/2019}
Devos, L., Meert, W., and Davis, J.
\newblock Fast gradient boosting decision trees with bit-level data structures.
\newblock In Brefeld, U., Fromont, {\'{E}}., Hotho, A., Knobbe, A.~J.,
  Maathuis, M.~H., and Robardet, C. (eds.), \emph{Machine Learning and
  Knowledge Discovery in Databases - European Conference, {ECML} {PKDD} 2019,
  W{\"{u}}rzburg, Germany, September 16-20, 2019, Proceedings, Part {I}},
  volume 11906 of \emph{Lecture Notes in Computer Science}, pp.\  590--606.
  Springer, 2019.
\newblock \doi{10.1007/978-3-030-46150-8\_35}.
\newblock URL \url{https://doi.org/10.1007/978-3-030-46150-8\_35}.

\bibitem[Fanelli et~al.(2013)Fanelli, Dantone, Gall, Fossati, and
  Van~Gool]{Fanelli/etal/2013}
Fanelli, G., Dantone, M., Gall, J., Fossati, A., and Van~Gool, L.
\newblock Random forests for real time 3d face analysis.
\newblock \emph{International journal of computer vision}, 2013.

\bibitem[Hakert et~al.()Hakert, Chen, and Chen]{Hakert/etal/2022}
Hakert, C., Chen, K., and Chen, J.
\newblock Flint: Exploiting floating point enabled integer arithmetic for
  efficient random forest inference.
\newblock abs/2209.04181.
\newblock \doi{10.48550/arXiv.2209.04181}.

\bibitem[Ke et~al.(2017)Ke, Meng, Finley, Wang, Chen, Ma, Ye, and
  Liu]{Ke/etal/2017}
Ke, G., Meng, Q., Finley, T., Wang, T., Chen, W., Ma, W., Ye, Q., and Liu, T.
\newblock Lightgbm: {A} highly efficient gradient boosting decision tree.
\newblock In Guyon, I., von Luxburg, U., Bengio, S., Wallach, H.~M., Fergus,
  R., Vishwanathan, S. V.~N., and Garnett, R. (eds.), \emph{Advances in Neural
  Information Processing Systems 30: Annual Conference on Neural Information
  Processing Systems 2017, December 4-9, 2017, Long Beach, CA, {USA}}, pp.\
  3146--3154, 2017.
\newblock URL
  \url{https://proceedings.neurips.cc/paper/2017/hash/6449f44a102fde848669bdd9eb6b76fa-Abstract.html}.

\bibitem[Kim et~al.(2010)Kim, Chhugani, Satish, Sedlar, Nguyen, Kaldewey, Lee,
  Brandt, and Dubey]{kim/etal/2010}
Kim, C., Chhugani, J., Satish, N., Sedlar, E., Nguyen, A., Kaldewey, T., Lee,
  V., Brandt, S., and Dubey, P.
\newblock {FAST}: {F}ast architecture sensitive tree search on modern {CPUs}
  and {GPUs}.
\newblock In \emph{Proceedings of the 2010 ACM SIGMOD International Conference
  on Management of data}, pp.\  339--350. ACM, 2010.

\bibitem[Kumar et~al.(2017)Kumar, Goyal, and Varma]{kumar/etal/2017}
Kumar, A., Goyal, S., and Varma, M.
\newblock {Resource-efficient machine learning in 2 KB RAM for the Internet of
  Things}.
\newblock In \emph{34th International Conference on Machine Learning, ICML
  2017}, 2017.
\newblock ISBN 9781510855144.

\bibitem[Lettich et~al.(2018)Lettich, Lucchese, Nardini, Orlando, Perego,
  Tonellotto, and Venturini]{lettich/etal/2018}
Lettich, F., Lucchese, C., Nardini, F.~M., Orlando, S., Perego, R., Tonellotto,
  N., and Venturini, R.
\newblock Parallel traversal of large ensembles of decision trees.
\newblock \emph{IEEE Transactions on Parallel and Distributed Systems},
  30\penalty0 (9):\penalty0 2075--2089, 2018.

\bibitem[Lucchese et~al.(2015)Lucchese, Nardini, Orlando, Perego, Tonellotto,
  and Venturini]{lucchese/etal/2015}
Lucchese, C., Nardini, F.~M., Orlando, S., Perego, R., Tonellotto, N., and
  Venturini, R.
\newblock Quickscorer: A fast algorithm to rank documents with additive
  ensembles of regression trees.
\newblock In \emph{Proceedings of the 38th International ACM SIGIR Conference
  on Research and Development in Information Retrieval}, pp.\  73--82, 2015.

\bibitem[Lucchese et~al.(2016)Lucchese, Nardini, Orlando, Perego, Tonellotto,
  and Venturini]{lucchese/etal/2016}
Lucchese, C., Nardini, F.~M., Orlando, S., Perego, R., Tonellotto, N., and
  Venturini, R.
\newblock Exploiting cpu simd extensions to speed-up document scoring with tree
  ensembles.
\newblock In \emph{Proceedings of the 39th International ACM SIGIR conference
  on Research and Development in Information Retrieval}, pp.\  833--836, 2016.

\bibitem[Lucchese et~al.(2018)Lucchese, Nardini, Orlando, Perego, Silvestri,
  and Trani]{lucchese/etal/2018}
Lucchese, C., Nardini, F.~M., Orlando, S., Perego, R., Silvestri, F., and
  Trani, S.
\newblock X-cleaver: Learning ranking ensembles by growing and pruning trees.
\newblock \emph{ACM Transactions on Intelligent Systems and Technology (TIST)},
  9\penalty0 (6):\penalty0 1--26, 2018.

\bibitem[Marín et~al.(2013)Marín, Vázquez, López, Amores, and
  Leibe]{Marin/etal/2013}
Marín, J., Vázquez, D., López, A.~M., Amores, J., and Leibe, B.
\newblock Random forests of local experts for pedestrian detection.
\newblock In \emph{2013 IEEE International Conference on Computer Vision},
  2013.
\newblock \doi{10.1109/ICCV.2013.322}.

\bibitem[Nakandala et~al.(2020)Nakandala, Saur, Yu, Karanasos, Curino, Weimer,
  and Interlandi]{nakandala/etal/2020}
Nakandala, S., Saur, K., Yu, G.-I., Karanasos, K., Curino, C., Weimer, M., and
  Interlandi, M.
\newblock A tensor compiler for unified machine learning prediction serving.
\newblock In \emph{14th $\{$USENIX$\}$ Symposium on Operating Systems Design
  and Implementation ($\{$OSDI$\}$ 20)}, pp.\  899--917, 2020.

\bibitem[Pedregosa et~al.(2011)Pedregosa, Varoquaux, Gramfort, Michel, Thirion,
  Grisel, Blondel, Prettenhofer, Weiss, Dubourg, et~al.]{pedregosa2011scikit}
Pedregosa, F., Varoquaux, G., Gramfort, A., Michel, V., Thirion, B., Grisel,
  O., Blondel, M., Prettenhofer, P., Weiss, R., Dubourg, V., et~al.
\newblock Scikit-learn: Machine learning in python.
\newblock \emph{the Journal of machine Learning research}, 12:\penalty0
  2825--2830, 2011.

\bibitem[Prokhorenkova et~al.(2018)Prokhorenkova, Gusev, Vorobev, Dorogush, and
  Gulin]{Prokhorenkova/etal/2018}
Prokhorenkova, L.~O., Gusev, G., Vorobev, A., Dorogush, A.~V., and Gulin, A.
\newblock Catboost: unbiased boosting with categorical features.
\newblock In Bengio, S., Wallach, H.~M., Larochelle, H., Grauman, K.,
  Cesa{-}Bianchi, N., and Garnett, R. (eds.), \emph{Advances in Neural
  Information Processing Systems 31: Annual Conference on Neural Information
  Processing Systems 2018, NeurIPS 2018, December 3-8, 2018, Montr{\'{e}}al,
  Canada}, pp.\  6639--6649, 2018.
\newblock URL
  \url{https://proceedings.neurips.cc/paper/2018/hash/14491b756b3a51daac41c24863285549-Abstract.html}.

\bibitem[Saki et~al.(2016)Saki, Sehgal, Panahi, and
  Kehtarnavaz]{Saki/etal/2016}
Saki, F., Sehgal, A., Panahi, I., and Kehtarnavaz, N.
\newblock Smartphone-based real-time classification of noise signals using
  subband features and random forest classifier.
\newblock In \emph{2016 IEEE International Conference on Acoustics, Speech and
  Signal Processing (ICASSP)}, pp.\  2204--2208, 2016.
\newblock \doi{10.1109/ICASSP.2016.7472068}.

\bibitem[Sharp(2008)]{sharp/2008}
Sharp, T.
\newblock Implementing decision trees and forests on a gpu.
\newblock In \emph{European conference on computer vision}, pp.\  595--608.
  Springer, 2008.

\bibitem[Shotton et~al.(2013)Shotton, Sharp, Kohli, Nowozin, Winn, and
  Criminisi]{shotton/etal/2013}
Shotton, J., Sharp, T., Kohli, P., Nowozin, S., Winn, J., and Criminisi, A.
\newblock Decision jungles: Compact and rich models for classification.
\newblock In \emph{Proceedings of the 26th International Conference on Neural
  Information Processing Systems}, 2013.

\bibitem[Summers et~al.()Summers, Guglielmo, Duarte, Harris, Hoang, Jindariani,
  Kreinar, Loncar, Ngadiuba, Pierini, Rankin, Tran, and Wu]{Summers/etal/2020}
Summers, S., Guglielmo, G.~D., Duarte, J.~M., Harris, P.~C., Hoang, D.,
  Jindariani, S., Kreinar, E., Loncar, V., Ngadiuba, J., Pierini, M., Rankin,
  D.~S., Tran, N., and Wu, Z.
\newblock Fast inference of boosted decision trees in fpgas for particle
  physics.
\newblock abs/2002.02534.
\newblock URL \url{https://arxiv.org/abs/2002.02534}.

\bibitem[Yayla et~al.(2019)Yayla, Toma, Chen, Lenssen, Shpacovitch,
  Hergenr{\"o}der, Weichert, and Chen]{yayla/etal/2019}
Yayla, M., Toma, A., Chen, K.-H., Lenssen, J.~E., Shpacovitch, V.,
  Hergenr{\"o}der, R., Weichert, F., and Chen, J.-J.
\newblock Nanoparticle classification using frequency domain analysis on
  resource-limited platforms.
\newblock \emph{Sensors}, 19\penalty0 (19):\penalty0 4138, 2019.

\bibitem[Ye et~al.(2018)Ye, Zhou, Zou, Gao, and Zhang]{ye/etal/2018}
Ye, T., Zhou, H., Zou, W.~Y., Gao, B., and Zhang, R.
\newblock Rapidscorer: fast tree ensemble evaluation by maximizing compactness
  in data level parallelization.
\newblock In \emph{Proceedings of the 24th ACM SIGKDD International Conference
  on Knowledge Discovery \& Data Mining}, pp.\  941--950, 2018.

\end{thebibliography}
	\bibliographystyle{mlsys2023}

	
	
\end{document}